\documentclass[a4paper,twoside]{article}

\usepackage{epsfig}
\usepackage{subfigure}
\usepackage{calc}
\usepackage{amssymb}
\usepackage{amstext}
\usepackage{amsmath}
\usepackage{amsthm}
\usepackage{multicol}
\usepackage{pslatex}
\usepackage[]{algorithm2e}
\usepackage{url}
\usepackage{array,booktabs}

\usepackage{SCITEPRESS}     % Please add other packages that you may need BEFORE the SCITEPRESS.sty package.
\usepackage{fancyhdr}
\begin{document}

\title{MASK: A flexible framework to facilitate de-identification of clinical texts}

\author{\authorname{Nikola Milosevic\sup{1}, and Gangamma Kalappa\sup{2}, and Hesam Dadafarin\sup{4}, and Mahmoud Azimaee\sup{2}, and Goran Nenadic\sup{1,3}}
\affiliation{\sup{1}School of Computer Science, University of Manchester, Manchester, United Kingdom}
\affiliation{\sup{2}Institute for Clinical Evaluative Studies, Toronto, Ontario, Canada}
\affiliation{\sup{3}Health e-Research Center, University of Manchester, Manchester, United Kingdom}
\affiliation{\sup{4}Evenset, Toronto, Ontario, Canada}
\email{\{nikola.milosevic,gnenadic\}@manchester.ac.uk}
}

\keywords{text mining, named entity recognition, de-identification, information extraction, natural language processing, clinical text }

\abstract{ 
Medical health records and clinical summaries contain a vast amount of important information in textual form that can help advancing research on treatments, drugs and public health. However, majority of these information is not shared because they contain private information about patients, their families or medical staff treating them. Regulations such as HIPPA in the US, PHIPPA in Canada and GDPR regulate the protection, processing and the distribution of this information.In case this information is de-identified and personal information are replaced or redacted, they could be distributed to the research community. In this paper, we present MASK, a software package that is designed to perform de-identification task. The software is able to perform named entity recognition using some of the state-of-the-art techniques and then mask or redact recognized entities. User is able to select named entity recognition algorithm (currently implemented are two versions of CRF-based techniques and BiLSTM-based neural network with pre-trained GLoVe and ELMo embeddings) and masking algorithm (e.g. shift dates, replace names/locations, totally redact entity).  
}

\onecolumn \maketitle \normalsize \vfill

\section{Introduction}

Medical research rely heavily on the information about patient treatments and their responses to the treatment \cite{schneeweiss2005review,calapodescu2017semi}. Most of the medical content, important for research, resides in the letters between physicians and in the clinical notes of clinicians since this is where they discussed their findings, explained ongoing treatments and outlined an overall view of the medical condition of the patient \cite{sweeney1996replacing}.  The adaptation of digital technologies in healthcare (such as electronic health records and digitalization of clinical documents) provide an interesting opportunity in clinical and public health research areas (for example, through large scale analysis of clinical documents). 

However, at the moment, most of the information about patients are held in isolated silos, in databases of individual medical institutions. Sharing the information to the research community would help realize the potential of research based on this data. However, institutions are facing challenges in sharing these information, out of which the main one is protecting privacy of patients and their staff \cite{calapodescu2017semi}. 

Privacy protection of patients is regulated by HIPPA (Health Insurance Portability and Accountability Act) in the US, PHIPA (Personal Health Information Protection Act) in Canada and GDPR (General Data Protection Regulation) in European Union. All these regulations mandate that medical records can be shared only de-identified. They also define what personal identifiable information (PII) and protected health information (PHI) are. While de-identification is not a hard problem in structural data (one need to redact certain fields), it is a challenge for textual information which may include some private information. However, it is common that textual information, which may be comments, therapy and its response description by clinician or hospital discharge summary contain very valuable information for the research. Human-based de-identification of efforts are very costly. For example, de-identification of 50,000 patient visit records in Medical Information Mart for Intensive Care-III (MIMIC-III) data set costed about \$500,000 and 5,000 hours of annotation work \cite{johnson2016mimic}. Despite the costs, it may still produces false negatives (one study found that recall of manual de-identification can range from 0.63 to 0.94, depending on clinician \cite{neamatullah2008automated}, while recall can be improved to the range of 0.998 - 1 for the teams of 3 annotators \cite{carrell2016juice}).    

While there have been attempts to tackle de-identification problem \cite{aberdeen2010mitre,sweeney1996replacing,stubbs2015automated}, even a number of shared tasks (e.g. i2b2), there are only few available tools for de-identification that can be used\footnote{\url{https://scrubber.nlm.nih.gov/}, }\footnote{\url{http://mist-deid.sourceforge.net/},}\footnote{\url{https://open.med.harvard.edu/wiki/display/SCRUBBER/Software}}. Most of the available tools are developed prior to 2014 and do not utilize novel advances in natural language processing. Also, the vast majority of tools that are available are performing just named entity recognition of the personal identifiable information (PPI). In order to truly de-identify a document, it is necessary to replace or redact personal identifiable information using some masking or redacting algorithm. None of the tools provide to the choice to utilize the best methodology for the given named entity, as they all utilize just a single method for extracting all named entities. 

In this paper, we present MASK framework, a python based framework for de-identification of clinical documents. It is a flexible framework, in which the user can select the algorithm for named entity recognition and masking of the identified entities. It already contains a number of implemented named entity recognition and masking algorithms. Also, it can be extended. Additionally, it can be used as a tool or as a python library in larger systems. The implemented algorithms can be trained or fine tuned on a new data sets, improving the performance of pre-trained models.

\section{Background}

The research on de-identification of medical records records started with Sweeney \cite{sweeney1996replacing}. He proposed a rule-based system called \textit{Scrub}. The system is based on a set of dictionaries of names, U.S. state abbreviations, and rules for each of the personal identifiable information. In the same year, a regulation that regulates sharing of personal identifiable information (HIPPA) in the U.S. have been passed. HIPPA defined 18 personal identifiable information that have to be de-identified in order to share medical information. 

The methodologies used for de-identification are usually either rule-based, machine learning based or a hybrid (combination of the the two). As we stated, early systems applied a rule-based approach \cite{sweeney1996replacing}. The latter approaches often utilized machine learning, most commonly CRF models.  

Important advances in developing de-identification methodologies were initialized by i2b2 challenges that held in 2006 and 2014 de-identification challenges \cite{uzuner2006i2b2,stubbs2015automated}. During these challenges de-identification data sets were published that are valuable resource reinforcing further research in the area. 

The tool called MIST (the MITRE Identification Scrubber Toolkit) utilized only conditional random fields in order to train named entity recogntion of PII/PHI \cite{aberdeen2010mitre}. MIST tool utilized only lexical features (e.g. word itself, previous and following words, whether the first letter is capital, etc.) for training the underlying CRF model. Using the MIST tool, it was possible to aid manual annotation and reduce annotation time from 3 minutes and 18 seconds to 1 minute and 20 seconds per note \cite{hanauer2013bootstrapping}. Other approaches based purely on CRF include \cite{phuong2016automatic,li2019efficient,henrikssona2018detecting,calapodescu2017semi}. While most of the approaches tackle de-identification in English language, Henriksonna et al. \cite{henrikssona2018detecting} are trying to perform de-identification of clinical notes in Swedish and Calapodescu et al. \cite{calapodescu2017semi} are tackling de-identification of medical documents in French.

The system called BoB (the best-of-breed) attempts to combine on UIMA architecture rules and CRF. It utilizes CTakes for several basic NLP tasks, such as sentence splitting, tokenization, chunking and part-of-speech tagging. Then it uses a combination of dictionary-based lookup rules and CRF. At the end the false positives are filtered using Support Vector machines \cite{ferrandez2012bob}. Similarly, Kim et al. \cite{kim2018ensemble} tested ensembles of various named entity recognition methods for NER, including CRF, LSTM-CRF, SVM, MIST, and rules and concluded that stacking methods improves the performance. 

Combination of rules, dictionaries and machine learning methods with two-pass recognition can yield good performance for finding entities that should be de-identified \cite{dehghan2015combining,yang2015automatic}. These approaches utilized a set of existing tools and vocabilaris to pre-process documents. Sets of generated features were used as input for CRF-based and rule-based named entity recognizer. In \cite{dehghan2015combining}, recognized entities were used as input in the second pass. Rules and dictionaries were used in the end to post-process the output, add missing entities and filter out false positives. 

Recently, there is emergence of neural networks approaches. For example, one approach is utilizing CBOW initialization and word representations using different kinds of recurrent neural networks \cite{yadav2016deep}. The more recent approach utilized bi-directional long-short term memory (LSTM) neural networks with combination of word (word2vec) and character-based embeddings \cite{dernoncourt2017identification}.

\section{Methodology}

We have developed a de-identification framework, called MASK, that can utilize multiple algorithms for both named entity recognition and masking. Accordingly, the MASK framework contains two components:
\begin{itemize}
\item Named entity recognition component
\item Masking component
\end{itemize}
The design of the framework allows named entity recognition and masking algorithms to be implemented and extended as a plugins. The framework is implemented in Python 3.7. It can be used as standalone application for de-identification or as a library with implemented de-identification function in a larger application.

\subsection{Named entity recognition component}

Named entity recognition is the crucial part of de-identification process, as entities that need to be replaced (de-identified), firstly need to be recognized. Mask provides a framework in which a set of named entity recognition algorithms can be trained and utilized with its plugin architecture. The architecture allows named entity recognition algorithms to be added by implementing an abstract class and five functions within it. The following functions need to be implemented: 

\begin{itemize}
\item \textbf{Initialization function} - in this function can be initialized variables and loaded necessary resources for new algorithm to work
\item \textbf{transform\_sequences} - this function takes a sequence tuples with a certain form (token, label) and transforms it to it to the sequence of features. It should return sequence of features and sequence of labels. 
\item \textbf{learn} - function for training of the algorithms (if needed, e.g. it is a machine learning algorithm). It takes a sequence of features and a sequence of labels and creates a model. 
\item \textbf{perform\_NER} - takes as the argument text and performs named entity recognition. Returns the sequence of tuples (token, predicted label). 
\item \textbf{evaluate} - the function for evaluating the algorithm that takes as input a sequence of feature and label instances. It should perform the algorithm on feature sequence and compare predicted labels to the actual ones. 
\end{itemize}

\begin{figure}[h!]
\centering
\includegraphics[width=80mm]{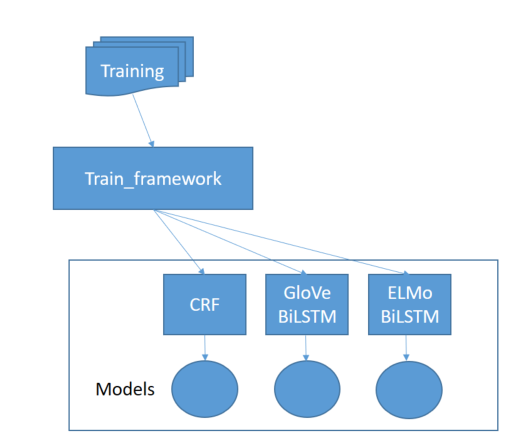}
\caption{MASK framework can train multiple approaches and algorithms on the data and then use the best performing model for each entity.}
\label{Figure Training}
\end{figure}

By using this architecture, multiple algorithms can be trained,  tested and compared using the same data set. Also, applying and fine-tuning algorithms for the new data sets is made simple. The architecture of the training system can be seen on Figure \ref{Figure Training}.

\begin{figure*}[ht!]
\centering
\includegraphics[width=120mm]{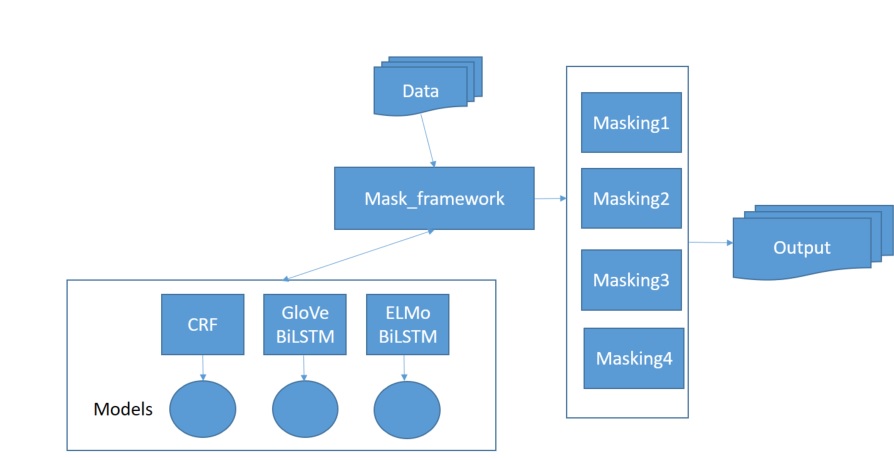}
\caption{MASK framework utilize pre-trained models and perform masking by selected masking algorithm}
\label{Figure Framework}
\end{figure*}

Once the algorithms are trained, the user can select the algorithm for each entity (either in configuration if it is used as a tool, or in code if used as a library). The named entity recognition is followed by masking algorithms that can be again selected by the user. Figure \ref{Figure Framework} presents components for applying MASK framework on a new data.

MASK framework so far implemented a several machine learning-based algorithms for named entity recognition. In the following subsections we explain the implemented methods. 

\subsubsection{CRF with lexical features only}

We trained conditional random fields algorithm on the set of training data from i2b2 2014 de-identification challenge. We employed some features engineering. The final set of features included the predicted word, whether the word is upper case, lower case, whether it starts with a capital letter, whether it contains only alpha-numeric characters, whether it contained only letter characters, and its shape. The shape of the word is generated as a string defining general shape of the token. Shape would be generated in a function that iterated each character of the token and generated stream by adding 'W' if the character was upper case, 'w' if it was lower case and 'd' if it was digit. Special characters remained. To capture context, we have used window of 4 tokens before and 4 tokens after the evaluated token. For tokens in the window, we have used the same features as for the evaluated token (whether it is upper-case, lower-case, starts with upper, numeric, alpha characters only, alpha-numeric, shape).

\subsubsection{CRF with dictionaries}

The second method, we evaluated uses as well conditional random fields as a machine learning algorithm and same lexical and morphological features as the previous method. However, in addition it used a set of dictionary matching features (whether a token is present in a given dictionary). The method used the following dictionaries: 
\begin{itemize}
    \item \textbf{Country dictionary} - lists the names of countries. The dictionary was generated by using GeoNames \footnote{\url{http://www.geonames.org/}}. The dictionary was manually filtered to contain only a single token names and synonyms of the countries. Multi-word countries were split into its tokens (e.g. Russian Federation - Russian, Federatation).
    \item \textbf{Cities dictionary} - list the names of cities that in GeoNames database have more than 500 inhabitants. Again, multi-token city names were split and each token was a separate entry in the dictionary (e.g. New York City - New, York, City). 
    \item \textbf{First names dictionary} - This dictionary was generated using the list of the most common baby male and female names in Ontario \footnote{\url{https://data.ontario.ca/dataset/ontario-top-baby-names-male}} \footnote{\url{https://data.ontario.ca/dataset/ontario-top-baby-names-female}}. By merging male and female lists, we have generated an terminology containing about 2000 names. 
    \item \textbf{Last name dictionary} - Contains a list of last names compiled from the publicly available data \footnote{\url{https://github.com/smashew/NameDatabases/tree/master/NamesDatabases}}. The available list contained surname from various cultures and countries. We have used a list US surnames that contains about 88700 instances. 
\end{itemize}

The dictionary features were flags indicating whether analysed word or words in the context window contain any of the tokens from the given dictionary. 

\subsubsection{BiLSTM using GLoVe embeddings}

Additionally to approaches using CRFs, we have developed two approaches using recurrent neural networks and word embeddings. The first of these approaches, used GLoVe (Global Vector) word embeddings \cite{pennington2014glove}. We used GLoVe embeddings trained on common crawl data containing 840 billion tokens, 2.2 million unique tokens in vocabulary, and 300-dimensional vectors. The vectors were input to a two layer bidirectional Long-Short Term Memory (BiLSTM) network, with the final dense layer with one output neuron for each class.  The first layer contains 150 LSTM units and the dropout is set to 0.3, and recurrent dropout is set to 0.6. The second layer contains 60 neuron with dropout of 0.2 and recurrent dropout of 0.5. The model was trained for 20 epochs.

\subsubsection{BiLSTM using ELMo embeddings}

The seconds approach used different embeddings (ELMo) and different architectural configuration of BiLSTMs. We have used ELMo embeddings \cite{Peters:2018} trained on common crawl data
 with embedding size of 1024\footnote{\url{https://tfhub.dev/google/elmo/2}}. 
 
 The neural network again contained two layers of bidirectional Long-Short Term Memory layers, both containing 512 neurons, 0.2 dropout and recurrent dropout. However, these two layers had a residual connection \cite{tran2017named} with the output layer, both connecting to it. These networks have additional spacial shortcut path between lower and higher levels of BiLSTMs and allow all learned information in all layers to be utilized for the final decision \cite{kim2017residual}. Again, the output layer was a dense layer with softmax activation function and number of output neurons equal to the number of classes. Similarly to the GLoVE model, the training was performed for 20 epochs. 
 
\subsection{Masking}

Once entities that make a record identifiable, they can be masked or redacted. Document where personal identifiable information are masked or redacted can be considered de-identified. MASK framework allows three operations:
\begin{itemize}
    \item \textbf{Redact} - redacts completely the entity. Redact operation replaces the entity with the fixed string.
    \item \textbf{Mask} - masks and replaces entity with more generic concept. For example, the address can be replaced with the city area, age or dates can be shifted, etc. There can be multiple masking functions for each entity that is being masked. 
    \item \textbf{Keep} - keeps the identified entity as is. This can be used in cases when certain entities are not sensitive or when the system is used for named entity recognition. 
\end{itemize}

We propose a plug-in based architecture, where user can select the masking algorithm from the library of implemented methods. Also, one can create and add to the library additional methods for masking by just following simple guidelines and implementing abstract masking class with two methods (initialization and application methods).

For certain research and use cases it is important that the document retains the structure as original. Also, documents should be de-identified in the way that certain entities (such as phone numbers, ids, names, dates) retain the original format. Therefore, we have developed a set of masking algorithms that replace identified entities with entities that do not belong the given individual or randomize and replace certain characters of the string entity (e.g. ID, phone number). 

At the moment of writing this paper, the following masking methods have been implemented: 
\begin{itemize}
    \item Masking names randomly from the list of names.
    \item Shifting dates by the set number of days.
    \item Masking profession with the random profession from the list of professions.
    \item Masking healthcare identifier by randomizing certain set of number within it.
    \item Masking phone number by randomizing several digits.
    \item Masking zip code by replacing last three characters.
\end{itemize}

New masking functions will continue to be implemented in the future. 

\section{Results}

We have trained and evaluated data on the i2b2 dataset\footnote{\url{https://portal.dbmi.hms.harvard.edu/}}. Also, we present experiences on applying data for de-identification of the data from clinics and laboratories in Ontario, maintained by Istitute for Clinical Evaluative Studies (ICES). 

\subsection{Data}

The i2b2 2014 data set contained in total 790  annotated clinical naratives.  The records were pulled from the Research Patient Data Repository of Partners Healthcare and de-identified, so the identifiable information were replaced with realistic fake information. The data contained patients diagnosed with diabetes. The data contained labelled following categories:
\begin{itemize}
    \item Name
    \item Profession
    \item Location
    \item Age
    \item Date
    \item Contact
    \item IDs
\end{itemize}
The distribution of annotated classes in the dataset can be seen in \cite{stubbs2015annotating}. 

We have used only this set, by training on 80\% of it and evaluating on the remaining 20\%. This would allow us to compare our system with the systems developed during the shared task. 

The i2b2 data were used as a primary training and evaluation data. We have also applied and tested methodology on ICES premises on the real data, but for this we only present qualitative experiences, while on the i2b2 we present both qualitative and quantitative evaluation. 

\subsection{Evaluation of named entity recognition}

We first compare the approaches for named entity recognition that we have developed on i2b2 data set. As mentioned before the models were trained on 80\% of data and tested on the remaining 20\%. The results of the evaluation are presented in Table \ref{table:perfModels}. 

\begin{table}[htbp]
\centering
\begin{tabular}{ lrrr }
  \hline
   \small{\textbf{Algorithm}} & \small{\textbf{Precision}} & \small{\textbf{Recall}} & \small{\textbf{F1-Score}}\\ \hline
  \small{CRF with lexical features} & \small{0.91} & \small{0.98} & \small{0.94} \\
   \small{CRF with dictionaries} & \small{0.92} & \small{0.97} & \small{0.95} \\
   \small{BiLSTM with GLoVe} & \small{0.92} & \small{0.92} & \small{0.92} \\
 \small{BiLSTM with ELMo} & \small{0.96} & \small{0.97} & \small{0.97} \\
\hline
\end{tabular} 
\caption{The overall performance by implemented algorithms, evaluated on i2b2 2014 data set}
\label{table:perfModels}
\end{table}

As it can be seen from the Table \ref{table:perfModels}, the best performing model was a deep neural network model using recurrent neural networks and ELMo embeddings. As ELMo embeddings and recurrent neural networks with residual connections have been in recent years proven as state of the art models in deep learning for sequential data, this could have been expected. The model outperforms all the systems that were presented on i2b2 challenge \cite{stubbs2015automated} as well as majority of the systems presented later \cite{kim2018ensemble,li2019efficient}. The only system that claimed slightly better results was \cite{dernoncourt2017identification}, claiming F1-score of 97.85\%, which is comparable and not statistically significant difference.

Conditional random fields (CRF) with the use of lexical and dictionary features perform slightly worse, but still beating systems evaluated on i2b2 challenge. The CRF system on CPU was significantly faster to both train and apply the named entity recognition. CRF with lexical features is as well able to perform quite well,  but there is an improvement that can be achieved with the use of crafted dictionaries of names, cities and countries. 

The worst performing system was biLSTM system with the use of GLoVe embeddings. This is most likely due to the use of common crawl word embeddings, rather than domain specific word embeddings. 

In the Table \ref{table:perfELMo}, we present the performance per entity of the best performing system - BiLSTM with residual connection and ELMo embeddings.

\begin{table}[htbp]
\centering
\begin{tabular}{ lrrrr }
  \hline
   \small{\textbf{Algorithm}} & \small{\textbf{Precision}} & \small{\textbf{Recall}} & \small{\textbf{F1-Score}}& \small{\textbf{Support}}\\ \hline
  \small{ID} & \small{0.83} & \small{1.00} & \small{0.90} & \small{85} \\
   \small{NAME} & \small{0.98} & \small{0.98} & \small{0.98} & \small{1100} \\
   \small{CONTACT} & \small{0.96} & \small{0.98} & \small{0.97}  & \small{56} \\
 \small{DATE} & \small{0.98} & \small{0.97} & \small{0.97}  & \small{919}\\
\small{AGE} & \small{0.96} & \small{0.96} & \small{0.96}  & \small{102}\\
\small{PROFESSION} & \small{0.84} & \small{0.89} & \small{0.86}  & \small{35}\\
\small{LOCATION} & \small{0.93} & \small{0.96} & \small{0.94}  & \small{580}\\
\small{PHI} & \small{0.00} & \small{0.00} & \small{0.00}  & \small{0}\\
\hline
\small{Overall} & \small{0.96} & \small{0.97} & \small{0.97}  & \small{2877}\\
\hline
\end{tabular} 
\caption{Performance by entity using BiLSTM and ELMo algorithm, evaluated on i2b2 2014 data set}
\label{table:perfELMo}
\end{table}

It can be noted that classes with larger amount of annotated instances perform better. The small classes would benefit from additional annotations. 

On the other hand, from practicality perspective, model that was using ELMo embedding was performing well and was generalizing quite well, even if applied on other domain. In order to fine-tune it for other domain, it needed relatively small training set. However, the downside of the ELMo model was that it was quite slow. Training on CPU could take about an hour per epoch. Applying the model could take up to 10 seconds per clinical narrative. 

The CRF models were fast compared to ELMo-based model. The complete training was taking up to 2 hours and it was taking less than half second per clinical narrative to apply named entity recognition. However, these models needed to be retrained completely for new domains. 

Therefore, regarding speed, generalizability and performance, one needs to evaluate the trade-off between the different models. 

%\subsection{Implemented masking functions}

\section{Conclusion}

In this paper, we presented MASK - a flexible framework for de-identification of clinical texts. The framework facilitates two step process of de-identification. Firstly, it recognizes named entities containing personal identifiable information. Once these information are identified, the framework allows different kinds of masking and redacting of identified information. The framework has flexible architecture, where methods for named entity recognition and masking can be implemented and added in form of plug-ins. Also, the framework is flexible in a sense that each named entity can be identified with different algorithm. This allow the user to set the algorithm that works best for the given named entity. Similarly, the framework allows the implementation of both rule-based and machine learning-based methods as plugins, including deep learning with modern word embeddings. Architecture also allows multiple input formats to be used for training and testing of machine learning algorithms for named entity recognition. This is possible by using or implementing different readers functions as plug-ins. 

We have implemented and evaluated several methodologies for named entity recognition and masking. The method we have implemented for named entity recognition included two conditional random field-based method, and two BiLSTM-based method with different word embeddings. The presented methods performed state-of-the-art performance on i2b2 dataset. We have also fine-tuned and applied the method on ICES premise for various domains and real use-cases of de-identification of clinical narratives. The experience with the use of the system was positive and it significantly improved and fastened the process of de-identification of records that were later shared and analysed for research on ICES premises. Similarly, masking algorithms were useful  for the given use-cases. 

Finally, we would like to build and encourage a community around the MASK. The MASK framework is open source and is available at \url{https://github.com/icescentral/MASK_public}. We encourage research community to use the frameworks as well as implement additional plugins for training input, NER algorithms and masking.

\section*{Acknowledgement}

This research is funded by HealTex and Health eResearch Centre (HeRC). The research and development was done in collaboration between the University of Manchester, Institute for Clinical Evaluative Sciences (ICES) and company Evenset. 
\bibliographystyle{acm}
\bibliography{TableMining}

\vfill
\end{document}